\documentclass[letterpaper]{article} 
\usepackage{aaai25}  
\usepackage{times}  
\usepackage{helvet}  
\usepackage{courier}  
\usepackage[hyphens]{url}  
\usepackage{graphicx} 
\usepackage{natbib}  
\usepackage{caption} 
\usepackage{algorithm}
\usepackage{algorithmic}
\usepackage{amsmath}
\usepackage{cleveref} 
\usepackage{booktabs} 
\usepackage{cleveref}

\usepackage{newfloat}
\usepackage{listings}
\DeclareCaptionStyle{ruled}{labelfont=normalfont,labelsep=colon,strut=off} 
\floatstyle{ruled}
\newfloat{listing}{tb}{lst}{}
\floatname{listing}{Listing}
\title{MaintaAvatar: A Maintainable Avatar Based on Neural Radiance Fields by Continual Learning}
\author{
    Shengbo Gu\textsuperscript{\rm 1,3},
    Yu-Kun Qiu\textsuperscript{\rm 1,3},
    Yu-Ming Tang\textsuperscript{\rm 1,3},
    Ancong Wu\textsuperscript{\rm 1,3}\thanks{Corresponding Authors.},
    {Wei-Shi Zheng\textsuperscript{\rm 1,2,3}}\footnotemark[1]
}
\affiliations{
    \textsuperscript{\rm 1}School of Computer Science and Engineering, Sun Yat-sen University, China\\
    \textsuperscript{\rm 2}Peng Cheng Laboratory, Shenzhen, China\\
    \textsuperscript{\rm 3}Key Laboratory of Machine Intelligence and Advanced Computing,
 Ministry of Education, China\\

    \{gushb3,qiuyk,tangym9\}@mail2.sysu.edu.cn, wuanc@mail.sysu.edu.cn, wszheng@ieee.org 
}

\usepackage{bibentry}

\begin{document}
\maketitle
\begin{abstract}
The generation of a virtual digital avatar is a crucial research topic in the field of computer vision. Many existing works utilize Neural Radiance Fields (NeRF) to address this issue and have achieved impressive results. However, previous works assume the images of the training person are available and fixed while the appearances and poses of a subject could constantly change and increase in real-world scenarios. 
How to update the human avatar but also maintain the ability to render the old appearance of the person is a practical challenge.
One trivial solution is to combine the existing virtual avatar models based on NeRF with continual learning methods.
However, there are some critical issues in this approach: learning new appearances and poses can cause the model to forget past information, which in turn leads to a degradation in the rendering quality of past appearances, especially color bleeding issues, and incorrect human body poses.
In this work, we propose a maintainable avatar (MaintaAvatar) based on neural radiance fields by continual learning, which resolves the issues by utilizing a Global-Local Joint Storage Module and a Pose Distillation Module. Overall, our model requires only limited data collection to quickly fine-tune the model while avoiding catastrophic forgetting, thus achieving a maintainable virtual avatar.
The experimental results validate the effectiveness of our MaintaAvatar model.
\end{abstract}

%

\section{Introduction}

Free-viewpoint rendering of 3D scenes has attracted significant academic and industrial attention, with effective solutions such as neural radiance fields (NeRF) and 3D Gaussian Splatting (3DGS) demonstrating impressive performance in this field.~\cite{nerf,barron2021mip,martin2018lookingood,martin2021nerf,wang2021ibrnet,zhang2020nerf++,kerbl20233d,qian20243dgs,li2024animatable,zhou2024feature,sun2024recent}.
Among them, human-specific methods \cite{neuralbody,liu2020neural,humannerf,sun2022human,hu2024gauhuman,pan2023transhuman,zhao2022humannerf,li2024animatable} model dynamic human bodies by establishing a deformable radiance field, achieving the driving and free-viewpoint rendering of dynamic human bodies.

\begin{figure}
  \centering
  \includegraphics[width=0.5\textwidth]{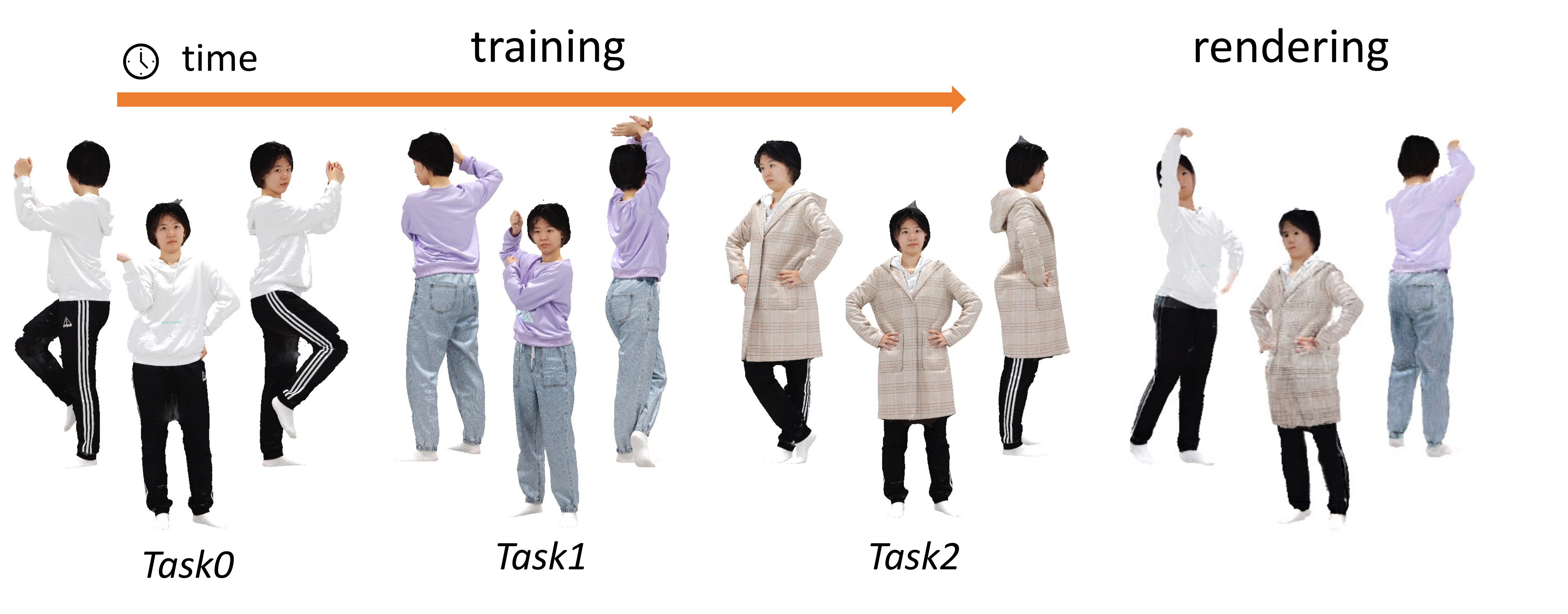} 
  \caption{In reality, a person's pose and appearance constantly update. Our maintainable avatar is designed to continuously learn from sequential data, enabling it to render any previously encountered viewpoint, pose, or appearance.}
  \label{fig:shouye}
\end{figure}

In this paper, we focus on creating a maintainable avatar that models a certain character with ever-changing poses and appearances (see Figure \ref{fig:shouye}).
In the real world, the poses and appearances of a person could be updated frequently, and consequently, the corresponding human avatar should also be updatable.
Employing conventional static human-nerf methods to track real-world trends will lead to a high demand for computation, training time, and extra storage room, let alone the need for abundant training samples from new styles and poses. 
We raise such a question:
\textit{Is it possible to maintain a dynamic human avatar using only a few shots from the updated character, with the ability to trace back to any point of time? }

The main challenge of such a task lies in the old avatar re-rendering with old poses and appearances, which is related to the well-known catastrophic forgetting \cite{french1999catastrophic,yang2023continual} of deep neural networks.
Specifically, if the model directly learns from new data, it tends to forget the information about past poses and appearances, resulting in poor rendering outcomes.
Further, we identify two major issues that employ the popular `replay' strategy that stores or generates (fake) old inputs for future task training~\cite{meil,cl-nerf,clnerf,shin2017continual} into a human avatar:
(1) The appearances between different tasks can interfere with each other, resulting in color bleeding and affecting the geometric rendering quality of each task. (2) During the process of continual learning of novel appearances and poses, the model may render incorrect poses in past tasks, consequently affecting the driving of the virtual avatar.

To address these problems, we propose two key solutions. First, the Global-Local Joint Storage Module stores the distinctions between global and local information of different appearances separately in global embeddings and Tri-planes. This allows the model to better differentiate the variations among different appearances. Second, the Pose Distillation Module extracts the pose information from past tasks as supervision signals for novel tasks, enabling the model to better retain the pose information of past tasks. 
Additionally, we create a pretrained model trained on multiple human bodies as the initialization for training, which can adjust the surface of the avatar model with only a few images, and quickly adapt to new human bodies.
We summarize our main contributions as follows:
\begin{itemize}
    \item 
    We study the problem of modeling a maintainable virtual avatar, which only needs a few shots to quickly update the character's appearance and also maintain the ability to render the old appearance. This maintainable virtual avatar can adapt in real-time to updates in the appearance of real-world individuals.
    \item 
    We analyze the two critical issues in the continual learning of maintainable virtual avatars. First,
    in order to prevent color bleeding between different appearances, we introduce the Global-Local Joint Storage Module that can precisely model the differences between various appearances.
    \item 
    Besides, to alleviate the information loss in the human pose, we introduce the Pose Distillation Module, which is capable of preserving the correct pose information of past appearances.

\end{itemize}
Experimental results in two datasets demonstrate the effectiveness of our model, achieving state-of-the-art performance.
\section{Related Work}
\label{sec:related}
\noindent\textbf{Neural Radiance Fields.}
NeRF \cite{nerf} is a widely acclaimed 3D technology that primarily focuses on synthesizing novel views. It represents a 3D environment as an implicit neural radiance field. Given a camera pose, it can generate images from any viewpoint using ray tracing. Specifically, NeRF \cite{nerf} utilizes a Multilayer Perceptron (MLP) \cite{multilayer} to map spatial coordinates and view directions to their corresponding colors and opacity. NeRF has sparked significant interest and extensive research across a range of domains, including applications in autonomous driving \cite{hu2024pc,chen2024s,cheng2023uc,feldmann2024nerfmentation}, controllable human avatars \cite{humannerf,personnerf,neuralbody,sun2022human,jiang2022neuman}, large urban scenes \cite{wang2023neural,xu2023grid,turki2023suds}, and text-driven 3D scene editing \cite{bao2023sine} and so on.

\noindent\textbf{Human-specific Neural Representation.}
Previous work \cite{humannerf,personnerf,zhao2022humannerf,hu2024gauhuman,hu2024gaussianavatar,neuralbody} expressed the human body as an implicit neural field, learning dynamic human figures by transforming bodies of different poses into a T-pose in a standard space. 
\cite{personnerf} learns human bodies with multiple appearances from unstructured data, addressing sparse observations in multi-outfit datasets by fusing appearance features via a shared network. However, it cannot create a virtual avatar that continually updates appearances while retaining the ability to render old ones. \cite{actorsnerf,mps,sherf,pan2023transhuman,kwon2023neural,zhao2022humannerf} propose a human body NeRF model that can quickly generalize to unseen human bodies in a few-shot setting without the need for continual learning. However, their performance is poor. 
\cite{liu2024humangaussian,kolotouros2024dreamhuman} generate virtual avatars through text guidance, and \cite{zhang2023text,mendiratta2023avatarstudio} further provide the capability of controllable editing, but they cannot create virtual avatars corresponding to real-world characters.
Recent works have leveraged Gaussian Splatting \cite{wu20234d,chen2023text,kerbl20233d,tang2023dreamgaussian} for human representation, achieving impressive results. \cite{hu2024gauhuman,qian20243dgs,hu2024gaussianavatar} further use it for fast training and real-time rendering.

\noindent\textbf{NeRF for Continual Learning.}
NeRF for continual learning is a new research hotspot. \cite{clnerf,meil,instant} propose a replay-based algorithm to establish continual learning NeRF, which can continuously learn new perspectives or dynamic scenes and achieve impressive results. \cite{cl-nerf} propose a lightweight expert adaptor to adapt to scene changes and a knowledge distillation learning objective to retain the invariant model. \cite{ILnerf} focusing on real-world scenarios with unknown camera poses. However, they cannot handle dynamic human bodies as they lack body modeling.

In contrast, MaintaAvatar is the first to propose maintaining dynamic virtual characters, enabling sequential learning of diverse poses, viewpoints, and appearances.
\label{sec:intro}

\begin{figure*}[t]
      \centering
      \includegraphics[width=1\linewidth]{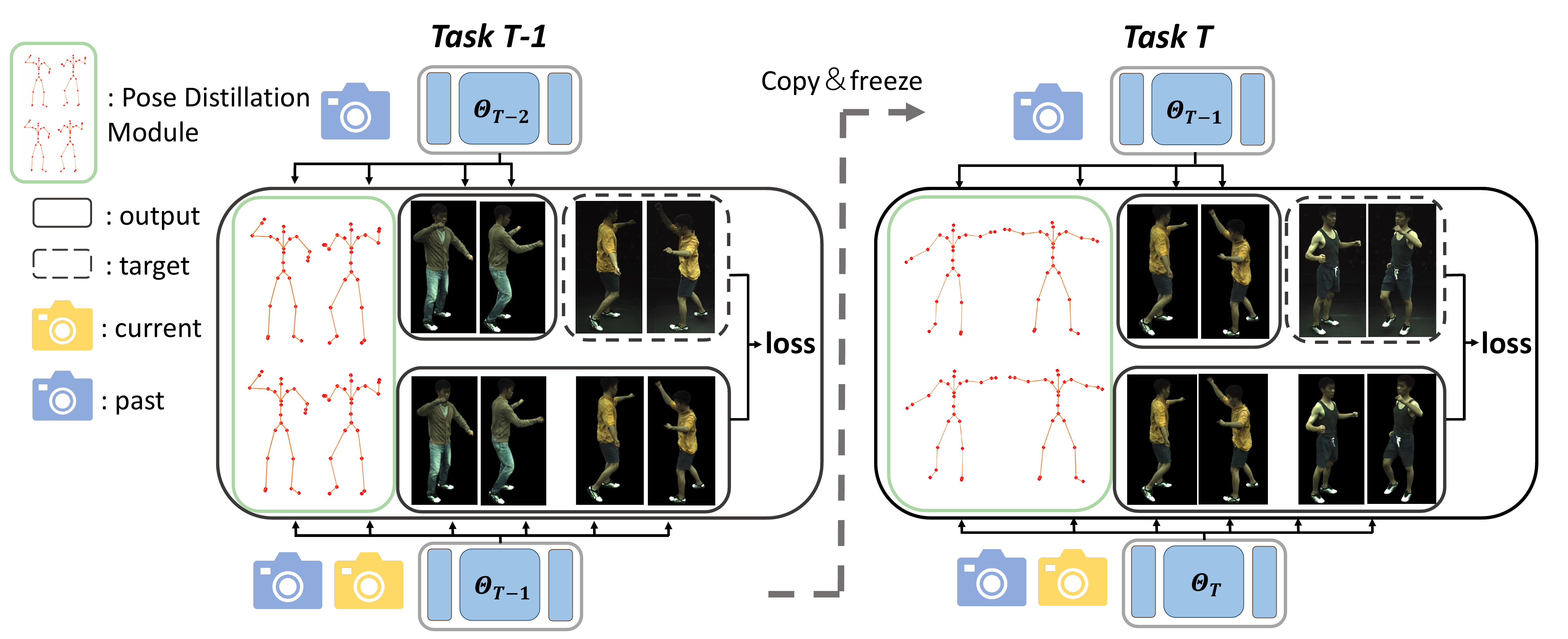}
      \caption{$\textbf  {MaintaAvatar Pipeline.}$ In this paper, we propose a continual learning strategy pipeline primarily based on the replay method. 
      During the training of \textit{Task T}, we replicate and freeze the network $ {\Theta}_{T-1}$ from the past \textit{Task T-1}. Given the camera parameters from \textit{Task T-1}, the network $\Theta_{ T-1}$ can generate corresponding patches and the residual human body pose of one randomly selected past appearance, which are utilized to supervise the training of \textit{Task T}. Simultaneously, $\Theta_T$ 
      is trained using images from the new appearance. In addition to image supervision, we incorporate a Pose Distillation Module to enhance the memory of past pose information, thereby improving rendering quality. Ultimately, our model is capable of continuously learning the novel appearance without forgetting past appearances.}
      \label{fig:continuelearing}
\end{figure*}

\section{MaintaAvatar}
\label{sec:network}
Our approach, MaintaAvatar, maintains a virtual avatar that can continuously learn novel poses and novel appearances without the need for retraining, saving both training and data collection costs. The pipeline is shown in Figure \ref{fig:continuelearing}. We introduce a deformation field to model the dynamic avatar (Section \ref{seq:deformation}) and employ a strategy based on generative replay to enable continual learning of the model, avoiding catastrophic forgetting (Section \ref{continual learning strategy}).   Building on this, we propose the Global-Local Joint Storage Module to model global and local appearance variations (Section \ref{sub:global-local}). 
Furthermore, we propose a Pose Distillation Module to mitigate the rendering of incorrect poses during the continual learning process (Section \ref{sub:pose}). 

\subsection{Preliminaries}

\noindent\textbf{Deformable NeRF Based on SMPL Model.}
\label{seq:deformation}
Following PersonNeRF \cite{personnerf}, we model NeRF \cite{nerf} as a deformable neural radiance field. Based on the observed SMPL model parameters, we can warp the canonical volume $\text{F}_c$ to the observed volume $\text{F}_o$ \cite{humannerf,personnerf}. 

Specifically, similar to PersonNeRF \cite{personnerf}, with SMPL model, we establish a rigid deformation field. By employing inverse linear skinning, we transform the vertices from the canonical volume $\text{F}_c$ into their corresponding positions in the observed volume $\text{F}_o$: 
\begin{equation}
\text{F}_\mathit{o} (\mathbf{x},\mathbf{p}) = \text{F}_c (\text{T}_{\mathit{skel}}(\mathbf{x},\mathbf{p}) ).
\end{equation}
where $\mathbf{p} = (\mathit J,\Omega)$ represents body pose, including joint positions $\mathit J$ and local joint rotations $\Omega$. $\mathbf{x}$ represents the coordinates of the sampling points. $\text{T}_{skel}$ is defined as rigid body motion, which maps an observed position to 
the canonical space:
\begin{equation}
\text{T}_{\mathit{skel}}(\mathbf{x},\mathbf{p}) = \sum^K_{i=1} \textit{w}^i_o(\mathbf{x})(\textit{R}_i \mathbf{x} + \textbf{t}_ \mathit{i}).
\end{equation}
where ($\textit{R}_i$,$\textbf{t}_ \mathit i$) is the mapping of the $i$-th bone from the observation volume to the canonical volume, which can be computed through body pose $\mathbf p$. $\textit{w}^i_o(\mathbf x)$ and $\textit{w}^i_c(\mathbf x)$ are the corresponding weight in the observed volume and the canonical volume \cite{humannerf,personnerf}, respectively. We derive the formula for computing $\textit{w}^i_o(\mathbf x)$ as:
\begin{equation}
\textit{w}^i_o(\mathbf x) = \dfrac{\textit{w}^i_c(\mathbf x)(\textit{R}_i \mathbf x + \textbf{t}_ \mathit i)}{\sum^K_{k=1} \textit{w}^k_c(\mathbf x)(\textit{R}_k \mathbf x + \textbf{t}_ \mathit k)}.
\end{equation}
To address the issue of inaccuracies in the body pose parameters from the dataset, 
some work utilizes a plug-and-play module $MLP_p$ \cite{humannerf,personnerf,hu2024gauhuman}. This module predicts the residual between the current body pose parameters and the true body pose parameters, correcting the body pose. As follow: 

\begin{equation}
{\Delta_\mathrm \Omega} (\mathbf p) = \text{MLP}_{\theta_{\rm pose}} (\rm \Omega).
\end{equation}
Then the pose $\mathbf p$ could be updated to:
\begin{equation}
\textit{P}_{\rm{pose}}(\mathbf{p}) = (\mathit{J},  {\Delta_ \mathrm \Omega} (\mathbf{p}) \otimes  \rm \Omega).
\end{equation}
\begin{figure*}[t]
  \centering
  \includegraphics[width=0.9\textwidth]{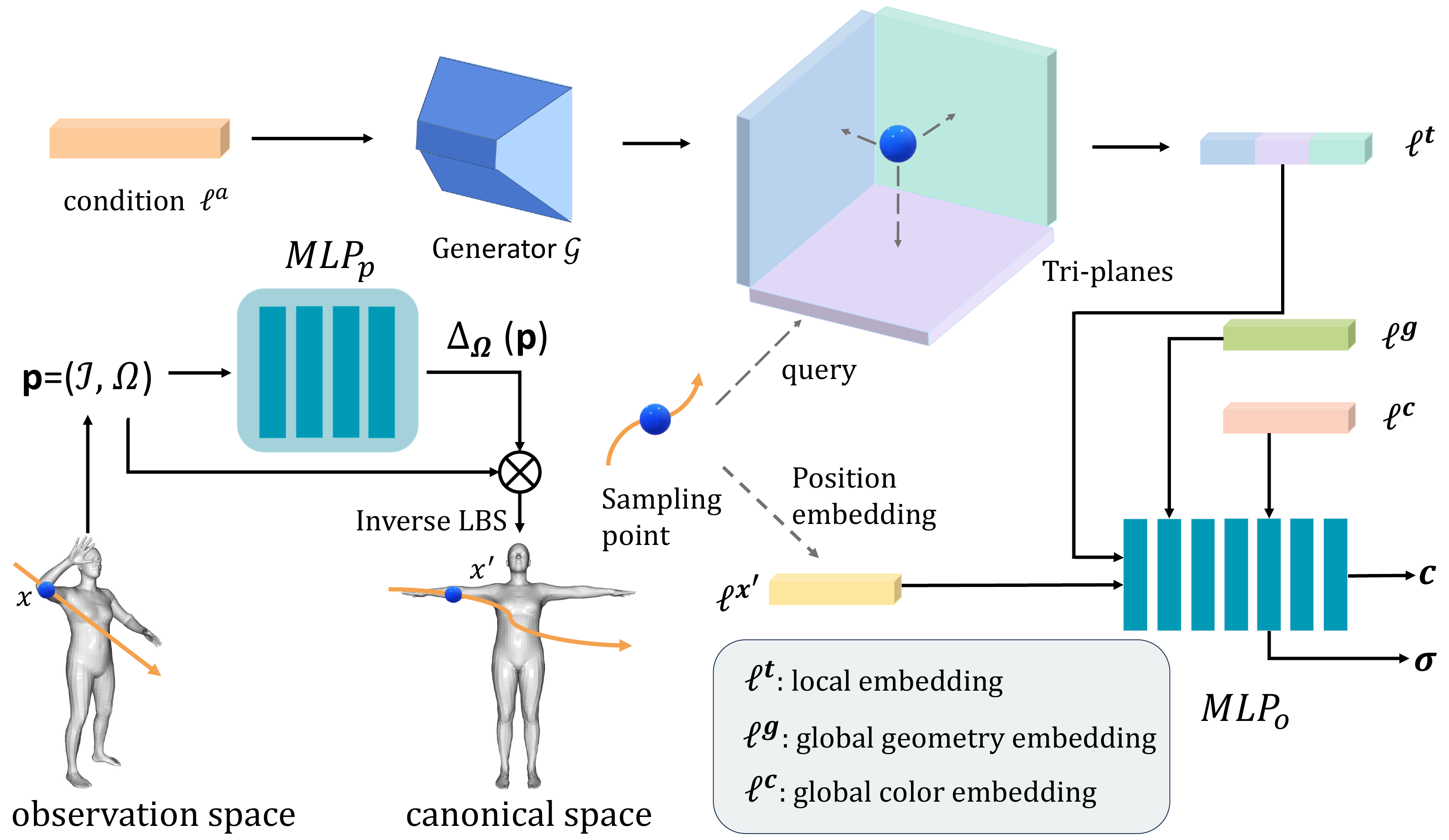} 
  \caption{$\textbf{Pipeline for MaintaAvatar Network}$ $\Theta$ $\textbf{Structure.}$ For any given human body pose, we utilize skeletal motion based on the SMPL model to transform the body from the observation space to the canonical space. Meanwhile, we employ a network $MLP_p$ to predict the residual $\Delta_\Omega(\mathbf{p})$ between the current pose parameters and the true pose parameters. Subsequently, our Global-Local Joint Storage Module generates Tri-plane-based local embedding and global embedding for each appearance. These embedding, along with coordinate embedding, are fed into $MLP_o$ to predict color and opacity.}
  \label{fig:network}

\end{figure*}
\noindent\textbf{Continual Learning for NeRF.}
\label{continual learning strategy}
The primary approach for continual learning for NeRF is to integrate a continual learning strategy with the NeRF network, parameterized by ${\Theta}$. And $\Theta_{ T}$ represents the network parameter of the current task, and $\Theta_{ T-1}$ represents the network structure of the past task. At every time step t of the continual NeRF: 

1. Freeze and copy the model $ {\Theta}_{T-1}$ from $\textit{Task T-1}$.

2. Given camera parameters, and pose parameters from the current task, current model $ {\Theta}_{T}$ firstly transforms the human body from the canonical volume $\text{F}_c$ to the observed volume $\text{F}_o$ and then utilizes an MLP network to render the color corresponding to each ray $\widehat C(\mathbf{r})$.

3. Utilizing the saved human body pose and camera parameters from past tasks, the network $ {\Theta}_{T-1}$ executes the process for step 2, and obtains the color supervision signal $\widetilde C(\mathbf{\bar r})$ of the past tasks.

4. The supervision signals $\widetilde C(\mathbf{\bar r})$ generated by model $\Theta_{ T-1}$ and the images obtained from the current task are used to supervise the training of model $\Theta_{ T}$.

\subsection{Global-Local Joint Storage Module}
\label{sub:global-local}
Most current researchs model dynamic scene variations using global geometry and color embeddings. However, our experiments show that this approach causes mutual interference between the novel and past appearances, leading to color blending (Figure \ref{fig:global}). 
Unlike the continual simple changes in dynamic scenes \cite{clnerf}, changes in human appearance usually involve drastic variations in color and geometry (Figure \ref{fig:shouye}). It is challenging to represent these significant spatial variations through global embeddings alone in continual learning, making global embeddings inadequate for accurately representing human appearance variations.

To address this challenge, we propose the Global-Local Joint Storage Module to model both global and local appearance variations. Global embeddings can represent the overall changes in human appearance, while local embeddings is used to represent the fine-grained variations based on these global changes. As shown in the Figure \ref{fig:network},
we randomly generate the global geometry embedding $\ell^g$ and color embedding $\ell^c$ to model global variations in appearance. On this basis, we introduce local embedding $\ell^t$ to model local variations superimposed on the global changes. 
To capture the local information of each appearance, we randomly generate a trainable condition embedding $\ell^a$ for each appearance, which is fed into the generator $\mathcal{G}$ to generate a Tri-plane. Then, for each sampling point on the ray, the local embeddings can be obtained by querying its position coordinates $x$ on the corresponding appearance Tri-plane.
\begin{equation}
\ell^t = q(\mathcal{G} (\ell^a) ,x)  
\end{equation}
Finally, we use the following formula to obtain the opacity and color of each sampling point:
\begin{equation}
{MLP_o}:(\gamma(x),\ell^g,\ell^c,\ell^t                             
             )\longrightarrow(c, \sigma).
\end{equation}
where $MLP_o$ is shown in Figure \ref{fig:network}. $\gamma(x)$, $\ell^g$, $\ell^c$, $\ell^t$ respectively represent the coordinate embedding, global geometry embedding, global color embedding and local embedding.

\begin{table*}[t]

\centering
\begin{tabular}{l|ccc|ccc}
\toprule
& \multicolumn{3}{c|}{\textbf{novel view}} & \multicolumn{3}{c}{\textbf{novel pose}} \\
\cmidrule{2-4} \cmidrule{5-7} 
\textbf{Method}& PSNR$\uparrow$ & SSIM$\uparrow$ & LPIPS$\downarrow$ & PSNR$\uparrow$ & SSIM$\uparrow$ & LPIPS$\downarrow$ \\
\midrule
Joint & 29.507 & 0.9652 & 34.579 &29.626&0.9702 &27.329\\
\midrule
$\text{CLNeRF}$ &12.3&0.927&152 &-&-&-                             \\ 
$\text{MEIL-NeRF}$ &25.810&0.9429&71.282 &25.902&0.9465&69.822                             \\ 
$\text{PersonNeRF}$ & 28.199 & 0.9626 & 44.354 &29.008&0.9665&36.724\\
$\text{PersonNeRF}_{CL}$ & 28.605 & 0.9632 & 38.826& 28.678 & 0.9665 & 32.889 \\
\midrule
 Ours & \textbf{29.495} & \textbf{0.9663} & \textbf{36.497} & \textbf{29.630} & \textbf{0.9702} & \textbf{30.166}\\

\bottomrule
\end{tabular}
\caption{The overall free-viewpoint and novel pose rendering experimental results on ZJU-MoCap  \cite{neuralbody}, which represent the evaluation metrics across all previously trained tasks using the weights trained on the final task. The best results are marked in bold.}
\label{tab:table1}
\end{table*}

\subsection{Pose Distillation Module}

\label{sub:pose}
Human pose is an important topic in the field of human body rendering. Inaccuracies in poses can lead to a significant degradation in rendering quality, resulting in unrealistic effects. $MLP_p$ is a plug-and-play method to correct pose errors in datasets, and is suitable for most human-specific neural representation. 
However, we found that current continual learning models for the human body may introduce incorrect poses from past tasks because $MLP_p$ overfits to learning new poses. Therefore, maintaining the accuracy of poses from past tasks during continual learning is a challenge to address. In this section, we propose a Pose Distillation Module. As shown in the Figure \ref{fig:continuelearing}. Similar to multi-layer distillation loss \cite{sun2019patient}, we distill the outputs of $MLP_p$ that perform pose correction.
\begin{equation}
\mathcal{L}_{\mathrm {POSE}} = \parallel \widehat \Delta_\Omega(\mathbf{p}) -   \widetilde \Delta_\Omega(\mathbf{p}) \parallel_2^2.
\label{eq:POSE}
\end{equation}
where $\widehat \Delta_\Omega(\mathbf{p})$  represents the output of the $MLP_p$ for the novel model and $ \widetilde \Delta_\Omega(\mathbf{p})$  represents the output of the $MLP_p$ for the previously frozen model, as shown in the Figure \ref{fig:network}. we calculate the loss using the L2 norm. In practice, our training process is divided into two phases. The first phase focuses on optimizing rendering quality, during which we do not activate the pose distillation loss function, as doing so could result in blurred rendering outcomes. However, in the second phase of training, which begins at $t_0$ as indicated in Equation \ref{eq:lambda}, we shift our focus to pose correction. In this phase, we activate the pose distillation loss function while freezing all components of the network except for the network $MLP_p$ and the color embedding $\ell^c$. This strategic shift allows the model to better capture and memorize the pose information from previous appearances. Additionally, the continual learning in color embedding will prompt the model to find a balance between the colors of the novel and past appearances.

\subsection{Optimization}
\subsubsection{Loss Function.}
The overall loss function is composed as follows:
\begin{equation}
\mathcal{L} = \lambda_1 \mathcal{L}_{\mathrm {CR}} + \lambda_p \mathcal{L}_{\mathrm {CL}} + \lambda_\beta \mathcal{L}_{\mathrm {POSE}}.
\end{equation}
$\mathcal{L}_{\mathrm {POSE}}$ is defined by Equation \ref{eq:POSE}. $\mathcal{L}_{\mathrm {CR}}$ represents the loss function for learning novel tasks, while $\mathcal{L}_{\mathrm {CL}}$ represents the loss function for memorizing past tasks. $\mathcal{L}_{\mathrm {CR}}$ and $\mathcal{L}_{\mathrm {CL}}$ are respectively defined by the following formulas:
\begin{equation}
\mathcal{L}_{\mathrm {CR}} = \sum_{\mathbf{r} \in \mathcal{R}}[\parallel  \widehat C(\mathbf{r}) -   C(\mathbf{r}) \parallel_2^2]   +  \lambda_2 LPIPS(\widehat C(\mathbf{r}), C(\mathbf{r})).
\label{eq:current loss}
\end{equation}

\begin{equation}
\mathcal{L}_{\mathrm {CL}} = \sum_{\mathbf{\bar r} \in \mathcal{\bar R}}[\parallel  \widehat C(\mathbf{\bar r}) -   \widetilde C(\mathbf{\bar r}) \parallel_2^2]   +  \lambda_2 LPIPS(\widehat C(\mathbf{\bar r}),\widetilde C(\mathbf{\bar r})).
\label{eq:incre loss}
\end{equation}
where $\mathbf{r}$ refer to the current rays of the set of rays $\mathcal{R}$ in each batch, while $\mathbf{\mathrm{\bar r}}$ refer to the past rays of the set of rays $\mathcal{\bar R} 
 $ in each batch. $\mathit C$, $\widehat{C}$ and $\widetilde C$  are the ground truth RGB colors, current network predicted RGB colors, and past network predicted RGB colors as supervisory signals. 
 
 We set $\lambda_2$ to 0.2. $\lambda_1$, $\lambda_p$ refer to weights for controlling the trade-off of current rays and past rays \cite{meil}. In which, $\lambda_1$ = 0.2 and $\lambda_p$ is defined by the following formula:

\begin{equation}
\label{eq:lambda}
\lambda_{p} = 
\begin{cases} 
\sin\left(-\frac{\pi}{2} + \frac{\pi(t-t_{\text{init}})}{t_{\text{max}}-t_{0}-t_{\text{init}}}\right) + 1 & \text{if } t < t_{\text{max}} - t_0 \\
1 & \text{otherwise}.
\end{cases}
\end{equation}
where $t_\text{init}$, $t_\text{max}$ respectively represent the initial and final iteration of the current task. $t_{0}$ represents the iteration to activate the Pose Distillation Module. For $\lambda_\beta$, if $t < t_{\text{max}} - t_0$, we set it to 0; otherwise, we set it to 800. It is worth noting that if $t<t_{max}-t_{0}$, we will freeze all networks except for network $MLP_p$ and the color embedding $\ell^c$.

\section{Experiments}
\subsection{Datasets}
\label{subsec:datasets}
Our model is evaluated on ZJU-MoCap \cite{neuralbody} \cite{neuralbody} and THuman2.0 dataset \cite{yu2021function4d}. 

\noindent\textbf{ZJU-MoCap \cite{neuralbody}.}
For ZJU-MoCap \cite{neuralbody}, we select subjects (377, 392, 393, 394) for our dataset, as they all feature the same individual in different sets of clothing. This dataset includes one camera assigned for training and the other 22 cameras for evaluation. For each task, we choose only five images with different viewpoints (ensuring a wide distribution of viewpoints as much as possible.) and different poses for training.

\noindent\textbf{THuman2.0 \cite{yu2021function4d}.}
For Thuman2.0 \cite{yu2021function4d}, we select subjects (262, 220, 207, 125) as the dataset. Thuman2.0 \cite{yu2021function4d} provides SMPL but does not offer rendered images or corresponding camera parameters. We use PyTorch3D to obtain rendered images from different viewpoints. To fully utilize the limited dataset, we render images from four viewpoints (0, 90, 180, 270 degrees) for each pose as the training set and render images from nine viewpoints (0, 40, 80, ..., 280, 320 degrees) for evaluation.

\begin{table*}[t]
\centering
\setlength{\tabcolsep}{3pt}
\begin{tabular}{lcccccccccccc}
\toprule
& \multicolumn{3}{c}{\textbf{subject 262}} & \multicolumn{3}{c}{\textbf{subject 220}} & \multicolumn{3}{c}{\textbf{subject 207}} & \multicolumn{3}{c}{\textbf{subject 125}} \\
\cmidrule(lr){2-4} \cmidrule(lr){5-7} \cmidrule(lr){8-10} \cmidrule(lr){11-13} 
\textbf{Method}& PSNR$\uparrow$ & SSIM$\uparrow$ & LPIPS$\downarrow$ & PSNR$\uparrow$ & SSIM$\uparrow$ & LPIPS$\downarrow$ & PSNR$\uparrow$ & SSIM$\uparrow$ & LPIPS$\downarrow$ & PSNR$\uparrow$ & SSIM$\uparrow$ & LPIPS$\downarrow$ \\
\midrule
$\text{CLNeRF}$ &12.9&0.94&236&12.9&0.938&235&14.2&0.913&123&10.8&0.907&224\\ 
$\text{MEIL-NeRF}$ &18.907&0.9344&78.161&17.142&0.9210&97.789&18.411&0.9409&72.456&15.484&0.9215&96.626\\ 
$\text{PersonNeRF}$  &20.094&0.9298&82.099&20.485&0.9338&78.517&21.560&0.9493&59.248&18.930&0.9390&71.145 \\
$\text{PersonNeRF}_{CL}$ & 20.905 & 0.9415 & 63.568 & 20.130 & 0.9365 & 72.227& 22.951 & 0.9624 & 45.814 & 19.448 & 0.9449 & 65.084 \\
 Our method & \textbf{22.316} & \textbf{0.9491} & \textbf{56.023} & \textbf{21.642} & \textbf{0.9449} & \textbf{64.173} & \textbf{23.998} & \textbf{0.9664} & \textbf{41.271} & \textbf{20.209} & \textbf{0.9488} & \textbf{60.884} \\
        
\bottomrule
\end{tabular}%
\caption{The overall free-viewpoint experimental results on Thuman2.0 \cite{yu2021function4d}. The best results are displayed in boldfaced font.}
\label{tabthuman2.0}
\end{table*}

\subsection{Implementation Details}
\label{subsec:setting}
The random seed is set to 42. The network $MLP_o$ and the $MLP_p$ have 8 and 4 layers respectively. The global color embedding $\ell^c$ (length 48), global geometry embedding $\ell^g$ (length 16), and condition embedding $\ell^a$ (length 16) are all optimized during training. The Tri-plane has dimensions of 3*512*512*8. We set the learning rates for the $MLP_o$ and both the $\ell^c$ and $\ell^g$ to $5\times10^{-4}$, and the rest to $5\times10^{-5}$. Adam is adopted as the optimizer. 
For the current task, we sample 6 patches of 32×32 size, whereas for past tasks, we sampled one patch of 64×64 size. 128 points are sampled from each ray. In the ZJU-MoCap dataset \cite{neuralbody}, each task is trained for 12,000 iterations, with the pose distillation loss (Equation \ref{eq:POSE}) inactive for the first 10,000 iterations and activated for the final 2,000. In contrast, in the Thuman2.0 dataset \cite{yu2021function4d}, tasks undergo 80,000 iterations, divided into two phases: the pose distillation loss remains inactive for the initial 70,000 iterations and becomes active for the last 10,000. To enable the model to quickly fine-tune to new human bodies, as well as to cope with a small amount of data, we constructed a pretrained model of other similarly sized human bodies as initialization to train the new human.

  
        

\subsection{Comparison with Other Methods}
\label{subsec:Results}

Up to now, there has been no ongoing research regarding the continual learning of multi-appearance human bodies. Therefore, we compare our model with $\mathrm{CLNeRF}$, $\mathrm {MEIL\text{-}NeRF}$, $\mathrm {PersonNeRF}$ and $\mathrm {PersonNeRF}_{CL}$, as shown in Table \ref{tab:table1}. Specifically, we respectively augment $\mathrm{CLNeRF}$ \cite{clnerf}, $\mathrm {MEIL\text{-}NeRF}$ \cite{meil} and $\mathrm {PersonNeRF}$ \cite{personnerf} with a pre-trained model as well as global geometry and color embeddings. Additionally, we apply the continual learning strategy of $\mathrm {MEIL\text{-}NeRF}$ to $\mathrm {PersonNeRF}$ as the $\mathrm {PersonNeRF}_{CL}$.
As a reference of performance upper bound, we report the results of training on the dataset of all tasks together denoted as ``Joint''. The evaluation metrics we use are PSNR \cite{hore2010image}, SSIM \cite{hore2010image}, and LPIPS. 

\noindent\textbf{Results on ZJU-MoCap \cite{neuralbody}.}
We first compare our method with others and ``Joint'' on ZJU-MoCap \cite{neuralbody}. The free-viewpoint and novel pose rendering results are shown in Table \ref{tab:table1}.
In all metrics, our method approaches the performance of ``Joint'' but does not fully reach it, while still outperforming all other methods.

\begin{figure}[t]
    \centering
    \includegraphics[width=1.0\linewidth]{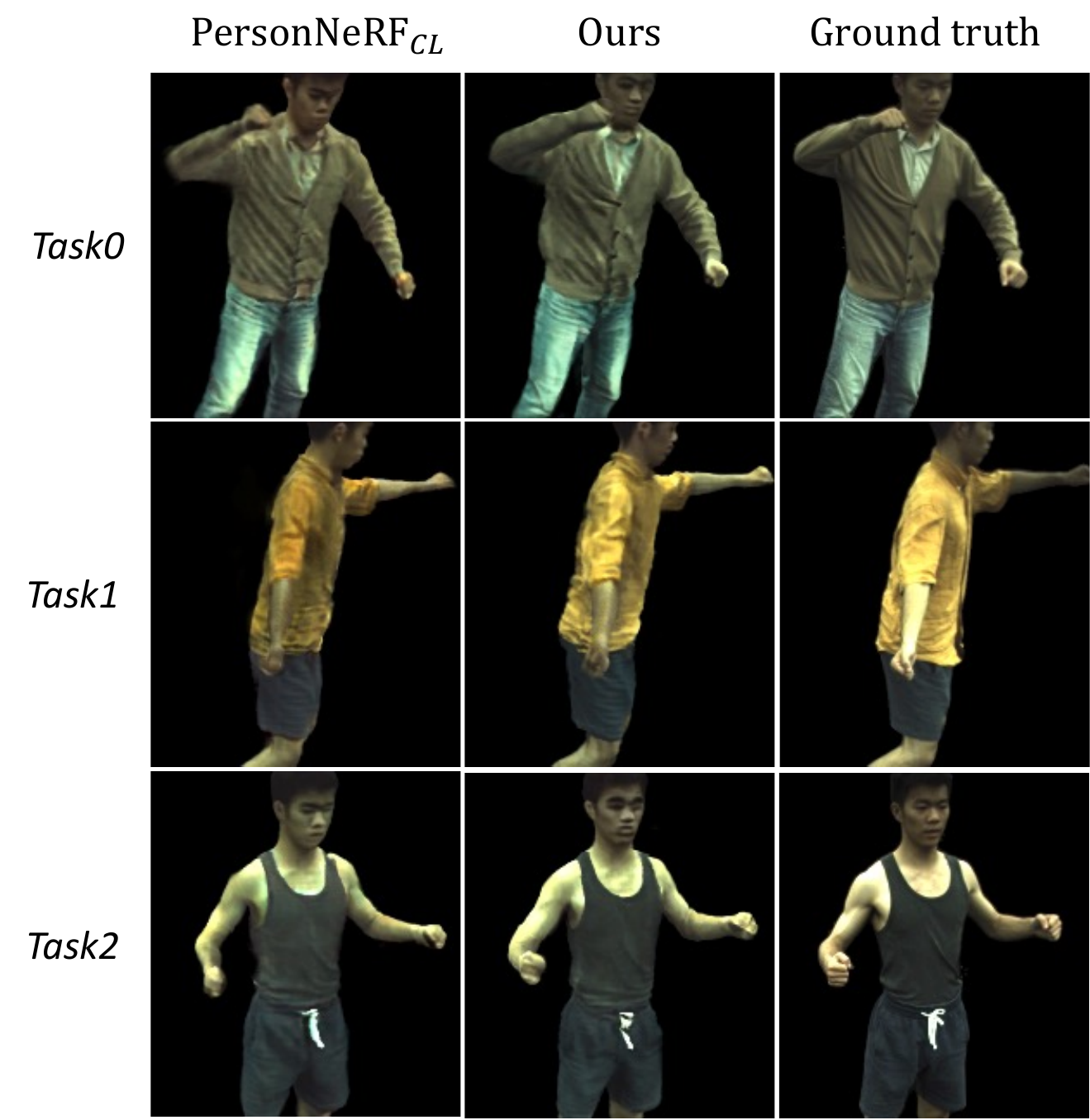}
    \caption{The visualization comparison results for the free-viewpoint rendering of our method and the $\mathrm {PersonNeRF}_{CL}$ on ZJU-MoCap \cite{neuralbody} for the past tasks. Our method demonstrates superior rendering quality, especially in terms of color and human pose.}
    \label{fig:global}
\end{figure}

The visualization results of free-viewpoint rendering on ZJU-MoCap \cite{neuralbody} are illustrated in Figure \ref{fig:global}. We use the model of final task ($Task3$). The $\mathrm {PersonNeRF}_{CL}$'s rendering of past tasks demonstrates issues with poor rendering quality and incorrect pose. Specifically, there are problems with missing details in color rendering and overall rendering inaccuracies in poses of the head, arms, and so on. Our proposed Global-Local Joint Storage Module and Pose Distillation Module effectively address these issues.




    
    

\begin{figure}[t]
  \centering
    \includegraphics[width=1\linewidth]{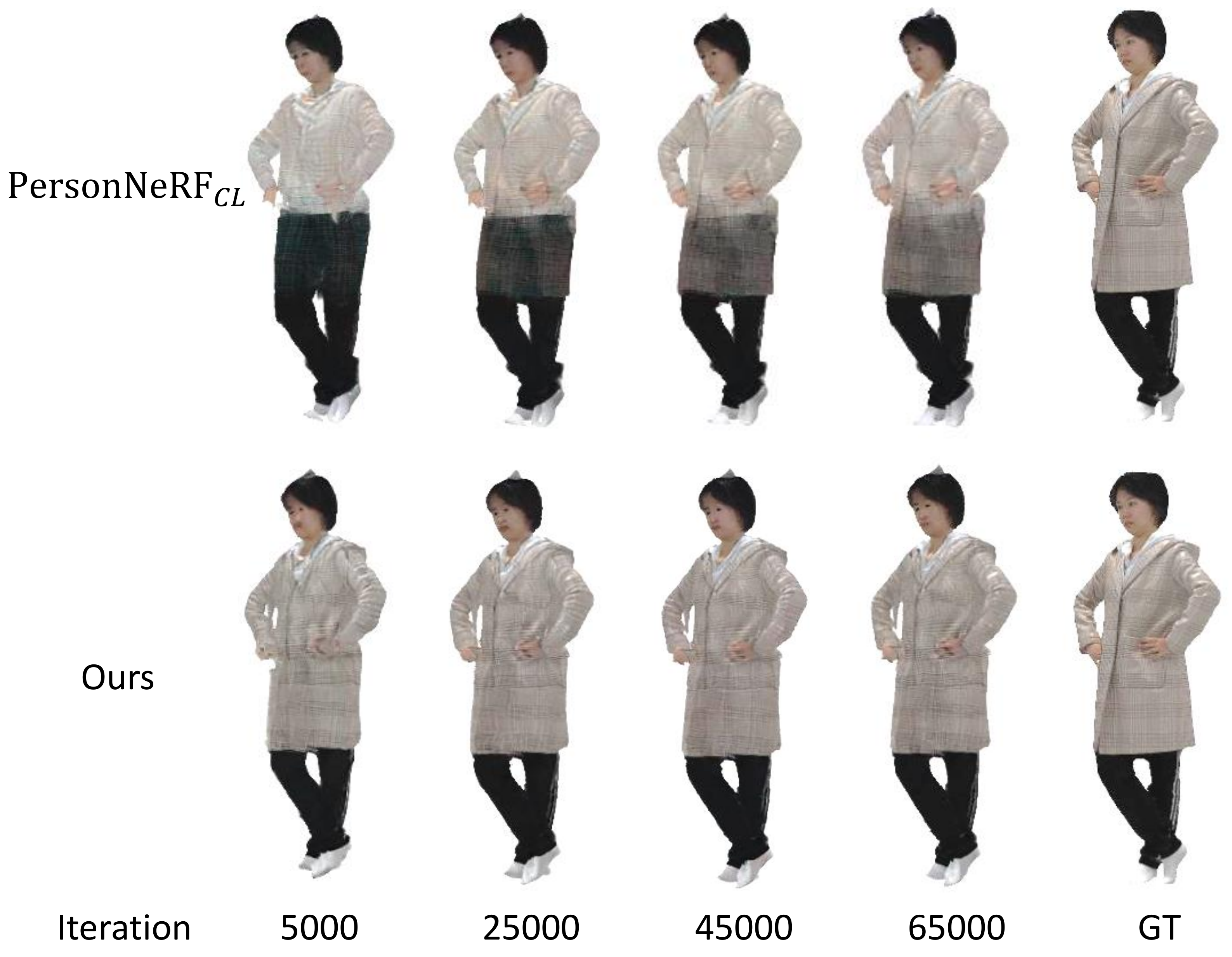}
    \caption{
    The Global-Local Joint Storage Module can help the model learn new appearances more quickly.}
    \label{fig:G-L消融Thu}
\end{figure}

\noindent\textbf{Results on Thuman2.0 \cite{yu2021function4d}.}
We further evaluated our model on Thuman2.0 \cite{yu2021function4d}. The results in Table \ref{tabthuman2.0} show that our model significantly outperforms the other models. At the same time, Figure \ref{fig:G-L消融Thu} demonstrates that our model with Global-Local Joint Storage Module can learn new appearances more quickly by better understanding the differences between different appearances.

\begin{table}[t]
\centering
\setlength{\tabcolsep}{3pt} 
\fontsize{10}{12}\selectfont
\begin{tabular}{l|ccc|ccc}
\toprule
& \multicolumn{3}{c|}{Novel View} & \multicolumn{3}{c}{Novel Pose} \\
\cmidrule(lr){2-4} \cmidrule(lr){5-7}
Method & PSNR & SSIM & LPIPS & PSNR & SSIM & LPIPS \\
\midrule
w/o pose & 29.21 & 0.9652 & 37.52 & 29.59 & 0.9698 & 30.83 \\
w/o G-L  & 28.89 & 0.9646 & 36.60 & 29.15 & 0.9686 & 30.99 \\
Full & \textbf{29.50} & \textbf{0.9663} & \textbf{36.50} & \textbf{29.63} & \textbf{0.9702} & \textbf{30.17} \\
\bottomrule
\end{tabular}
\caption{Ablations in ZJU-MoCap: ``w/o G-L'' omits the Global-Local Joint Storage Module, ``w/o pose'' omits the Pose Distillation Module.}
\label{tab:ablations}
\end{table}

\begin{figure}[t]
  \centering
    \includegraphics[width=\linewidth]{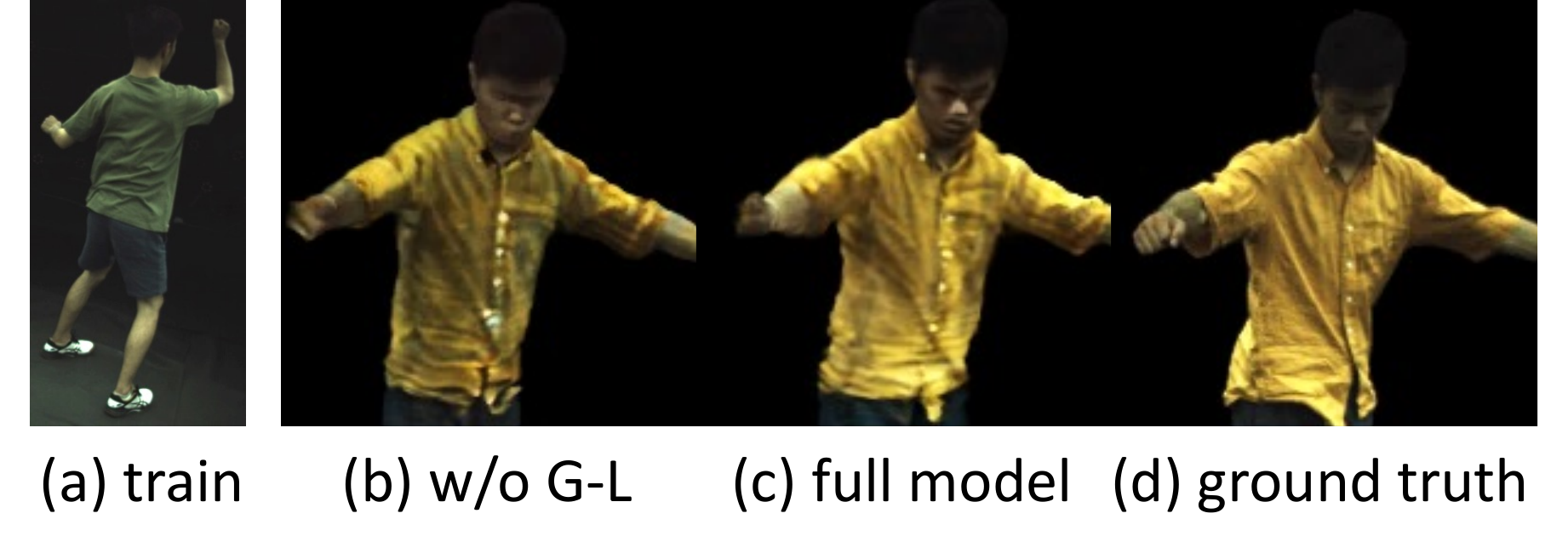}
    \caption{
    The rendering result for the past task (yellow apperance) after training on the current task (green apperance). The Global-Local Joint Storage Module effectively prevents color bleeding between different appearances.}
    \label{fig:G-L消融}
\end{figure}


    
   
   
\begin{figure}
    \centering
        \includegraphics[width=0.65\linewidth]{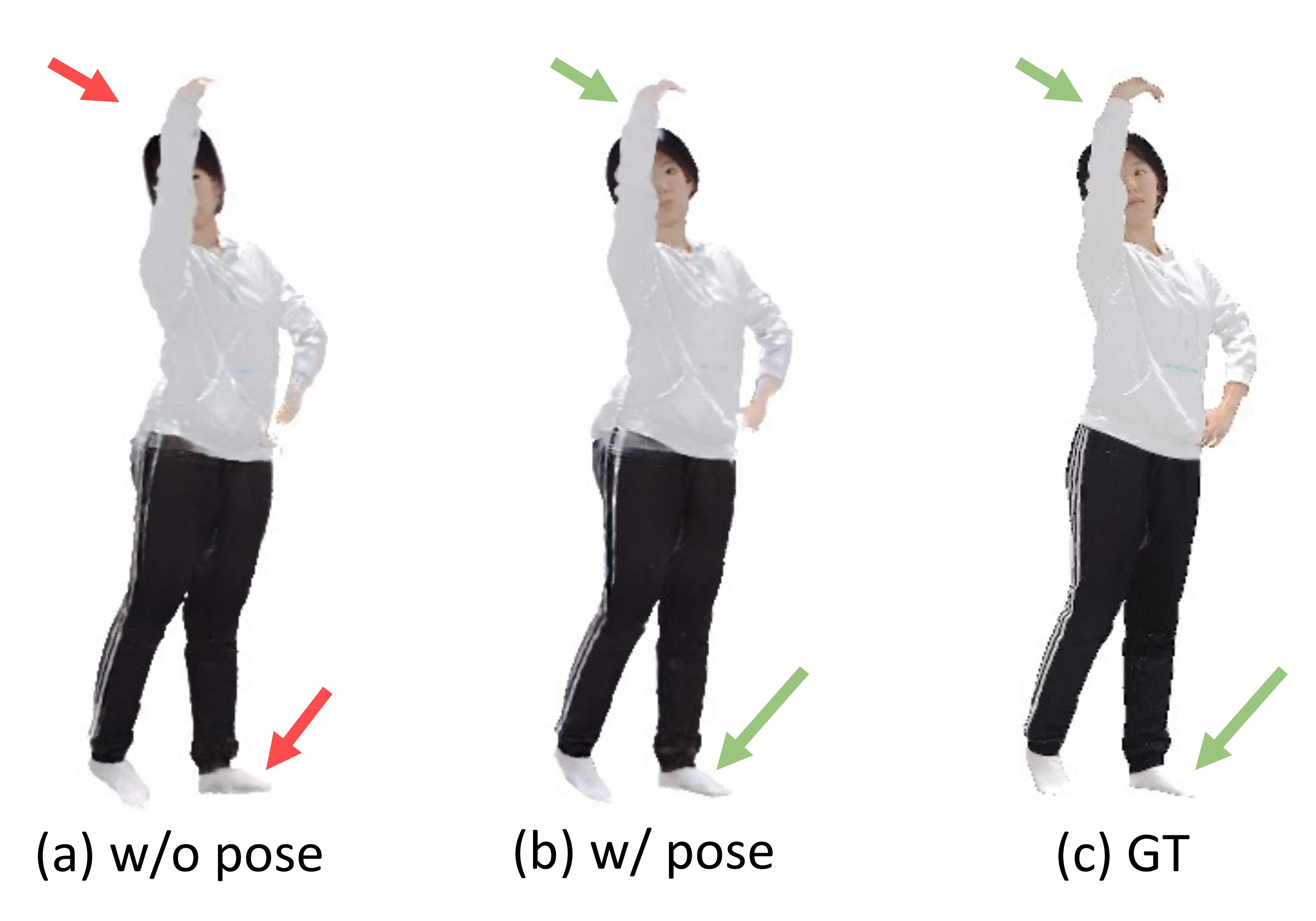}
        \caption{The pose distillation loss function improves the quality of novel view synthesis. The Pose Distillation Module is referred to as ``pose''.}
        \label{fig:image1}
    
    \end{figure}
    \begin{figure}
    \centering
        \includegraphics[width=0.65\linewidth]{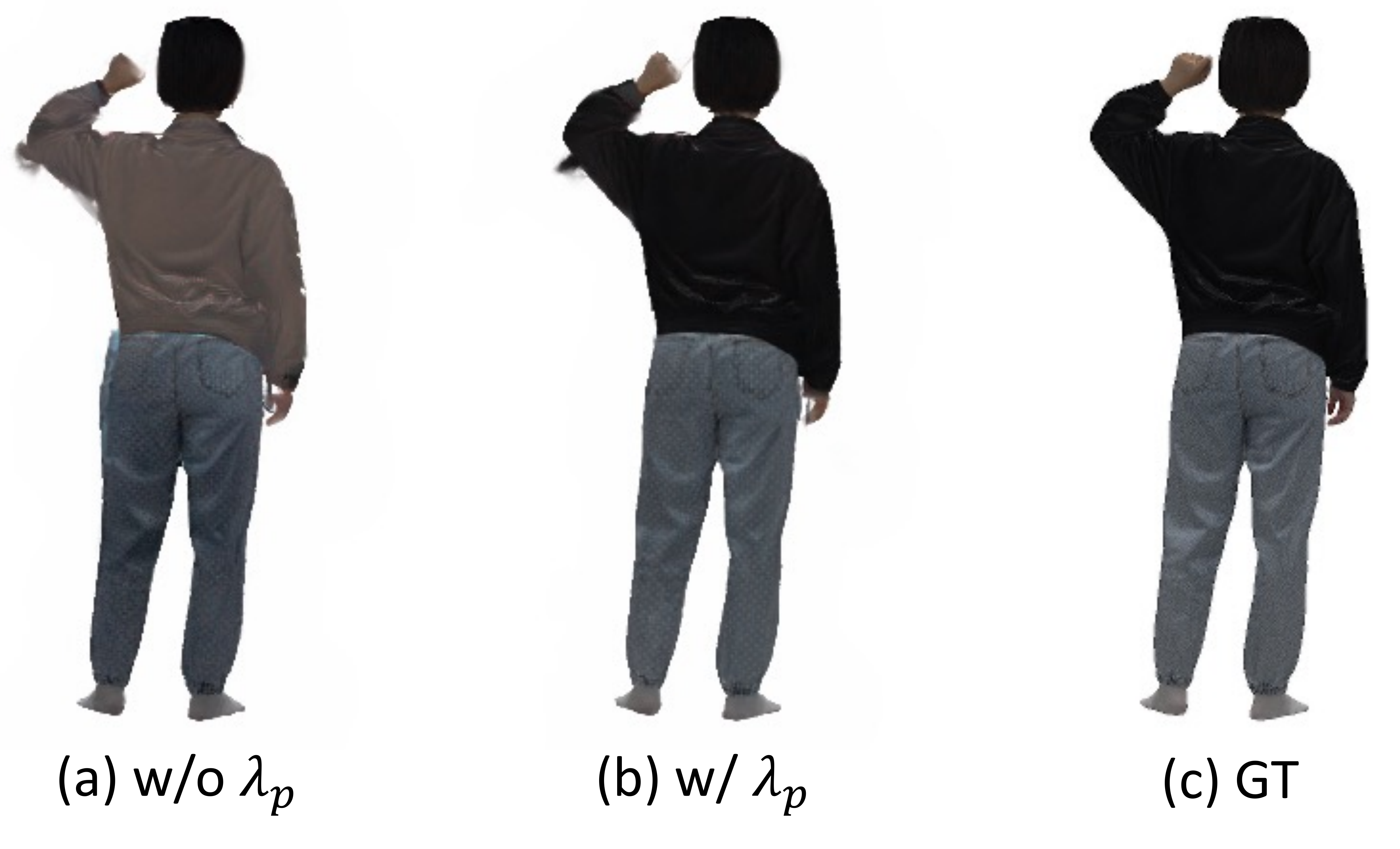}
        \caption{The hyperparameter $\lambda_p$ enables the model to find balance between current appearance and past appearance.}
        \label{fig:lambdap}
\end{figure}

\subsection{Ablation Studies}
\label{subsec:Ablations}

\noindent\textbf{Global-Local Joint Storage Module.}
``w/o G-L'' denotes the results without using the Global-Local Joint Storage Module. ``full model'' denotes our full model. Table \ref{tab:ablations} and Figure \ref{fig:G-L消融} demonstrate the rendering improvements brought by the Global-Local Joint Storage Module. We use the old model with green clothing to learn yellow clothing. Without using this module, the generated yellow appearance is influenced by the dark green appearance. Figure \ref{fig:G-L消融Thu} demonstrates the training speed improvements.



\noindent\textbf{Pose Distillation Module.} ``w/o pose'' denotes the results of removing the Pose Distillation Module from our model. The ablation study regarding pose distillation is shown in Figure \ref{fig:image1} and Table \ref{tab:ablations}. We present the visualization experimental results on Thuman2.0 \cite{yu2021function4d}. The Pose Distillation Module is capable of rendering a human figure with the correct pose more effectively.

\noindent\textbf{$\lambda_p$ Hyperparameter.}
The hyperparameter $\lambda_p$ controls the trade-off between current and past tasks, as demonstrated in the ablation experiment shown in Figure \ref{fig:lambdap}. $\lambda_p$ is defined as the Equation \ref{eq:lambda}. Without $\lambda_p$, the model struggles to balance the current task and past tasks, especially in cases where there is a significant color difference between current and past appearances.

\section{Limitations and Conclusion}
\noindent\textbf{Limitations.} Our method shows performance drops with significant clothing shape changes and struggles with pose generalization in complex poses due to limited exposure in the few-shot dataset.

\noindent\textbf{Conclusion.} MaintaAvatar is the first work to propose a maintainable virtual avatar to address the issue of continual changes in human appearance. We introduce two main components: a Global-Local Joint Storage Module to prevent color bleeding between different appearances and a Pose Distillation Module to solve the pose forgetting problem in continual learning. MaintaAvatar requires only a minimal set of training images to fine-tune to a novel appearance within a fixed, shorter training time, without forgetting the past appearances. 

\section*
{Acknowledgments}
This work was supported partially by NSFC (92470202, U21A20471), Guangdong NSF Project (No. 2023B1515040025).

\bibliography{aaai25}

\end{document}